\documentclass{article}
\usepackage{ijcai24}

\usepackage{times}
\usepackage{soul}
\usepackage{url}
\usepackage[hidelinks]{hyperref}
\usepackage[utf8]{inputenc}
\usepackage[small]{caption}
\usepackage{graphicx}
\usepackage{amsmath}
\usepackage{amsthm}
\usepackage{booktabs}
\usepackage[switch]{lineno}

\usepackage{algorithm}
\usepackage[utf8]{inputenc} %
\usepackage{booktabs}
\usepackage{amsfonts}       %
\usepackage{nicefrac}       %
\usepackage{microtype}      %
\usepackage[dvipsnames]{xcolor}         %
\usepackage{xspace,mfirstuc,tabulary}
\usepackage{enumerate}
\usepackage{enumitem}
\usepackage{multirow}
\usepackage{array}
\makeatletter
\newcolumntype{V}[1]{>{\topsep=0pt\@minipagetrue}p{#1}<{\vspace{-\baselineskip}}}
\newcolumntype{L}[1]{>{\raggedright\let\newline\\\arraybackslash\hspace{0pt}}m{#1}}
\newcolumntype{C}[1]{>{\centering\let\newline\\\arraybackslash\hspace{0pt}}m{#1}}
\newcolumntype{R}[1]{>{\raggedleft\let\newline\\\arraybackslash\hspace{0pt}}m{#1}}

\makeatother

\usepackage{comment}
\newcommand\scalemath[2]{\scalebox{#1}{\mbox{\ensuremath{\displaystyle #2}}}}
\newcommand{\myparagraph}[1]{\noindent\textbf{#1}}
\newcommand{\yes}{\textbf{\textcolor{teal}{Yes}}}
\newcommand{\teal}[1]{\textbf{\textcolor{teal}{#1}}}
\newcommand{\no}{\textcolor{Mahogany}{No}}

\usepackage[utf8]{inputenc}
\usepackage{subcaption}
\usepackage{amsmath}
\usepackage{algorithm}
\usepackage[noend]{algpseudocode}

\newcommand{\ra}{$\rightarrow$}
\usepackage{tabularx}
\newcommand{\citet}[1]{\citeauthor{#1}, \citeyear{#1}}

\newcommand{\gapxx}{\hspace*{4mm}}

\newcommand{\gapxxxxxx}{\hspace*{12mm}}

\newcommand{\gapxxxxxxxx}{\hspace*{16mm}}

\newcommand{\gapxxxxxxxxxx}{\hspace*{20mm}}

\definecolor{dyellow}{RGB}{216,182,98}
\definecolor{dgreen}{RGB}{132,178,108}
\definecolor{dpurple}{RGB}{139,86,162}
\definecolor{dblue}{RGB}{144,157,225}
\definecolor{dorange}{RGB}{207,158,56}
\definecolor{dblue2}{RGB}{7,95,198}

\usepackage{microtype}

\newcommand\sysname{\textsc{Nellie}}
\newcommand\nellie{\textsc{Nellie}}

\title{\sysname{}: A  Neuro-Symbolic Inference Engine \\ for Grounded, Compositional, and Explainable Reasoning}

\author {
    Nathaniel Weir\textsuperscript{\rm 1},
    Peter Clark\textsuperscript{\rm 2},
    \textnormal{and}
    Benjamin Van Durme\textsuperscript{\rm 1}
    \affiliations
    \textsuperscript{\rm 1}Johns Hopkins University, Baltimore, MD, USA\\
    \textsuperscript{\rm 2}Allen Institute for AI, Seattle, WA, USA\\
    \emails
    \{nweir, vandurme\}@jhu.edu, peterc@allenai.org
}

\begin{document}

\maketitle

\begin{abstract}

Our goal is to develop a modern approach to answering questions via systematic reasoning where answers are supported by human interpretable proof trees grounded in an NL corpus of facts.
Such a system would help alleviate the challenges of interpretability and hallucination with modern LMs, and the lack of grounding of current explanation methods (e.g., Chain-of-Thought). This paper proposes a new take on Prolog-based inference engines, where we replace handcrafted rules with a combination of neural language modeling, guided generation, and semiparametric dense retrieval. Our implementation, \sysname{}, is the first system to demonstrate fully interpretable, end-to-end grounded QA as entailment tree proof search, going beyond earlier work explaining known-to-be-true facts from text. In experiments, \sysname{} outperforms a similar-sized state-of-the-art reasoner
while producing knowledge-grounded explanations. 
We also find \sysname{} can exploit both semi-structured and NL text corpora to guide reasoning. 
Together these suggest a new way to jointly reap the benefits of both modern neural methods and traditional symbolic reasoning.

\end{abstract}
\section{Introduction}

Large language models (LLMs) are impressive at question-answering (QA), but it remains challenging to
understand how answers systematically follow from authoritative information.
This general opacity is a growing impediment to widespread use of LLMs, e.g., in
critical applications such as medicine, law, and hiring decisions, where interpretability and trust are paramount.
While there has been rapid progress in having LLMs generate systematic
explanations for their answers, e.g., Chain-of-Thought~\cite{wei-etal-2022-chain},
Entailer~\cite{tafjord-etal-2022-entailer}, or Maieutic Prompting \cite{jung-etal-2022-maieutic},
such explanations are not grounded in external facts and may include
hallucinations~\cite{ji-etal-2022-survey}.
Rather, what would be desirable - and what this work pursues - is a system
that systematically reasons over authoritative text: to support answers with human interpretable proof trees grounded in the text, while not requiring translation to an entirely symbolic formalism.

\begin{figure}[t!]
\includegraphics[width=\columnwidth,trim={2.5mm 2mm 2mm 2mm},clip]{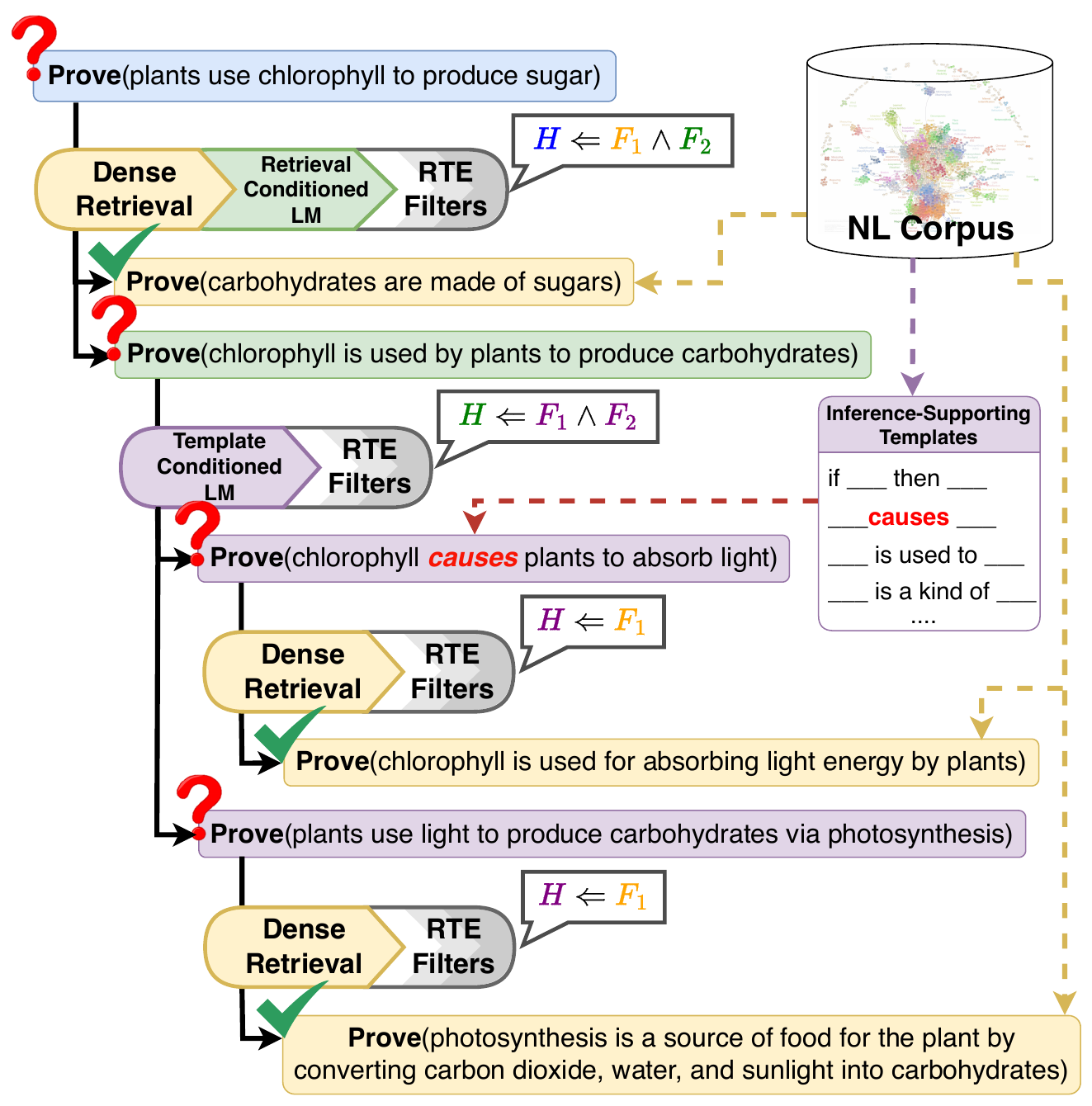}
\caption{
Given a \textcolor{dblue}{\textbf{query}}, \sysname{} performs a neuro-symbolic backward chaining search
for proof trees whose \textcolor{dyellow}{\textbf{leaves}} are grounded in a corpus of facts. 
It generates candidate decomposition rules guided by \textcolor{dgreen}{\textbf{retrieved
facts}} or \textcolor{dpurple}{\textbf{templates}}. Then, it recursively tries to prove rule conditions
via entailment from the corpus or further decomposition.
}

\label{fig:nellie_cartoon}
\end{figure}

Our approach is to revisit the behavior of
\textit{expert systems}~\cite{jackson-1986-introduction,metaxiotis-etal-2002-expert}.
Expert systems are appealing for their explainable behavior: decisions are made by
constructing a well-formed symbolic proof from explicit, formally represented knowledge
authored by a knowledge engineer in consultation with a domain expert.
However, as expert systems are known to be both expensive and brittle~\cite{musen-etal-1988-brittleness},
the AI community has turned to \textit{neuro-symbolic} mechanisms that
use large language models to reason over natural language (NL).
Reasoning over NL does not require a symbolic formalism
and allows the use of inferentially powerful LLMs, but
also loses the systematic, verifiable proofs that expert systems produced. 
This motivates our pursuit of a new way to jointly reap the benefits of both modern neural
methods and traditional reasoning.

We desire the following expert system-inspired desiderata:
\begin{enumerate}
    \item \textbf{Grounding} inferences fully and scalably in a corpus of knowledge from an authoritative human source.
    \item \textbf{Logical direction}, showing how a given hypothesis (and all intermediate inferences leading up to it) follows as the logical consequence of the underlying knowledge source
    \item \textbf{Competitive end-to-end QA performance} in a complex domain requiring diverse forms of reasoning.
\end{enumerate}

\begin{figure}[t!]
    \centering
    
  \includegraphics[width=.45\textwidth]{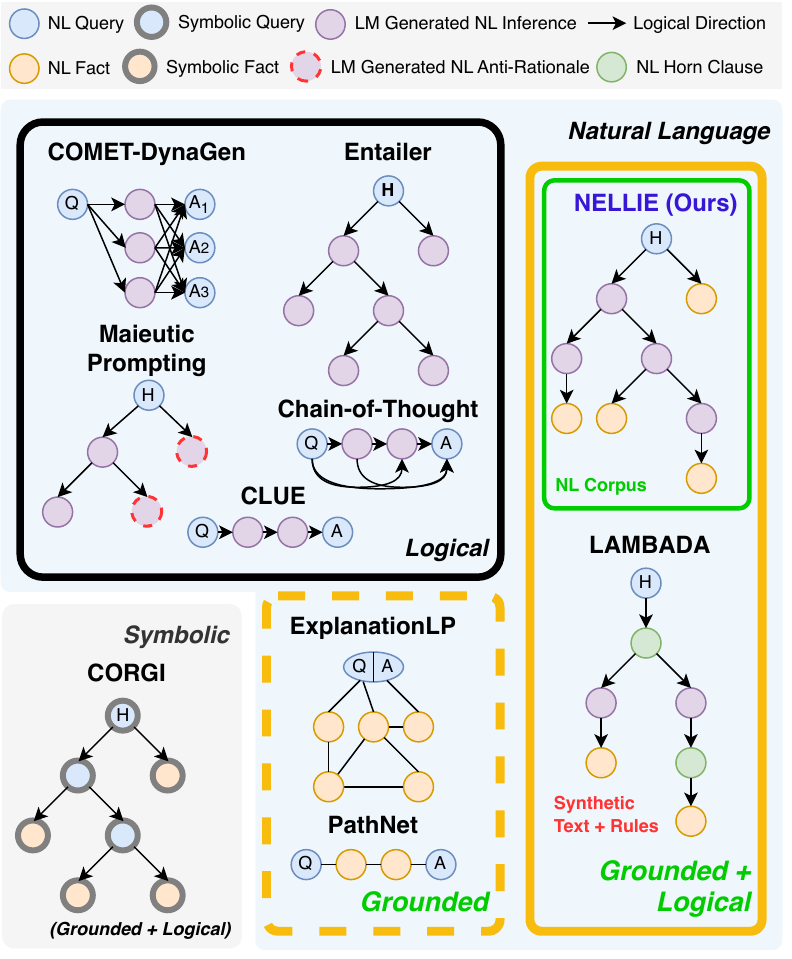}
    \caption{Comparison of approaches to neural XQA.
    Each approach leads to NL graphs in support of a \textcolor{dblue}{\textbf{query}}.
    Our proposal is to produce logically directed explanations containing \textcolor{dpurple}{\textbf{model-generated intermediate steps}} while grounding a tree in \textcolor{dorange}{\textbf{verified facts}} without relying on \textcolor{dgreen}{\textbf{handwritten horn clauses}}.}
    \label{fig:approaches}
\end{figure}

As illustrated in \autoref{fig:approaches}, various eXplainable QA (XQA) methods accomplish 2 of these criteria, but not all 3.  Some methods, like Chain-of-Thought, generate inference chains or logically structured graphs from an LLM without grounding in verified knowledge.
Others, like ExplanationLP~\cite{thayaparan-etal-2021-explainable}, compose graphs of grounded facts, but do not show logical direction.
LAMBADA~\cite{kazemi-etal-2023-lambada} achieves both direction and grounding but is limited to simple domains with provided NL Horn rules and sets of 1-2 dozen facts. 
In contrast, we desire a system that handles a larger corpus (10K statements) and doesn't need handcrafted rules.

To achieve all three desiderata, we reuse the general inference framework of an expert system, but replace handcrafted rules with a combination of
neural language modeling, guided generation, and semiparametric dense retrieval. We demonstrate this in a system called \sysname{}, the 
\textbf{Ne}uro-Symbolic 
\textbf{L}arge 
\textbf{L}M
\textbf{I}nference 
\textbf{E}ngine. \sysname{} leverages LLMs as \textit{proposal functions} in the search of proof trees showing how a conclusion follows via entailment from an external corpus of NL statements. 
The ``symbols'' that our neuro-symbolic engine reasons over are free-form NL sentences.
Rather than require a knowledge engineer to carefully write hundreds of inference rules as in the classical setting,
\sysname{} employs an LLM as a \textit{dynamic rule generator}~[DRG; \citet{kalyanpur-etal-2021-braid}]
to \textit{generate} (rather than retrieve) candidate rules that decompose a hypothesis into subqueries that,
if themselves proved recursively, will prove that hypothesis via composition (Figure \ref{fig:nellie_cartoon}).
In this way, \sysname{} can answer the question posed by the classical expert system: ``Does this statement systematically follow from my knowledge, or not?".

\sysname{} is built upon a backward chaining symbolic theorem prover written in Prolog, implemented using a few \emph{meta-rules} specifying how inference should proceed. 
These include checking for
entailment of a hypothesis against a retrieved fact 
or decomposing it into a conjunction of subqueries to recursively prove.
To treat NL sentences as if they were symbols in a purely symbolic search algorithm, we use 
a natural language inference-based form of \textit{weak unification}~\cite{sessa-2002-approximate,weber-etal-2019-nlprolog} between the hypothesis and
a corpus fact. This produces interpretable proofs similar to the compositional entailment trees of
prior work (e.g., EntailmentBank~\cite{dalvi-etal-2021-explaining}) 
while tackling the additional challenge of QA.

\sysname{} is designed to search across hundreds of trees whose leaves come from one of two types of knowledge: (a) a corpus of semi-structured text statements,
such as NL renditions of database entries, or (b) a corpus of free-form NL sentences. 
Many \textit{correct} proofs might exist for a given hypothesis, but much fewer will be \textit{fully groundable} in the provided corpus, making the search harder than for ungrounded alternatives like Entailer~\cite{tafjord-etal-2022-entailer} or COMET-Dynagen~\cite{bosselut-etal-2021-dynamic}.
To improve grounding, we introduce two guiding methods to boost the likelihood of generating rules
whose conditions match against the available text: 
(1) For applications where a semi-structured corpus
is available, \sysname{} leverages this structure via templates to help steer rule generation towards
syntax that is more likely to match corpus entries.
(2) For both free-form and semi-structured corpora, \sysname{} retrieves and conditions on statements to help steer generation towards trees grounded in them.
Ablation
experiments show these together and individually improve \sysname{}'s reasoning.

\sysname{} expands upon methods that ground known-to-be-true hypotheses to provided facts~[SCSEARCH; \citet{bostrom-etal-2022-natural}],
extending the paradigm 
to perform QA. This involves searching for and comparing trees for conflicting
answer options. %
Experiments find \sysname{} outperforms a similar-sized state-of-the-art reasoner, Entailer, while producing compositional trees showing how decisions are grounded in corpus facts. 
Our contributions are thus:
\begin{enumerate}
\item An architecture for systematic reasoning from a corpus of textual knowledge to
answer questions. This can be viewed as a modern approach to expert system inference,
but without requiring a formal knowledge representation.
\item An implementation, \sysname{},\footnote{Code found at \url{https://github.com/JHU-CLSP/NELLIE}.
} that outperforms a similar-sized SOTA reasoner (Entailer-3B)
while producing grounded trees. To our knowledge, this is the
first system to perform grounded XQA as NL entailment tree search.
\end{enumerate}

\section{Related Work}
\paragraph{Theorem Proving over Language} A long-standing approach to reasoning over NL is to project it into a symbolic form, such as for QA~\cite{green-61-BASEBALL,zelle-96} or entailment~\cite{bos-markert-2005-recognising}. Provided the translation from NL to symbolic representation is accurate, one can leverage fast and scalable solutions for discrete symbolic inference~\cite{riazanov2002design,kautz1992planning,kautz1999unifying}.
However, reliably translating broad-domain NL into an adequately expressive formalism for reasoning is a challenge~\cite{Schubert_2015}, though some have found success using LLMs to perform this semantic parsing task~\cite{wong-etal-2023-word,lyu-etal-2023-faithful,olausson-etal-2023-linc,pan-etal-2023-logic,ye-etal-2023-satlm}.

Recent work explores methods of handling NL without parsing it to another formalism.
Some use LMs to generate proof steps in mathematical theorem proving~\cite{polu-sutskever-2020-GenerativeLM,welleck-etal-2022-naturalprover}. 
Variants of neural theorem provers~[NTPs;~\citet{rocktaschel-riedel-2017-end}] such as NLProlog~\cite{weber-etal-2019-nlprolog} reconcile NL facts with symbolic reasoning by learning embeddings for the facts and symbols in a theory, then using weak unification to backward chain.
\citet{kalyanpur-etal-2021-braid} inject neural reasoning into a Boxer/Prolog-based symbolic reasoner via a special LM-calling predicate. \citet{arabshahi-etal-2021-corgi} combine handwritten symbolic rules with neural symbol embeddings to classify conversational intents.
Other work explores using LMs to emulate stepwise~\cite{tafjord-etal-2021-proofwriter,kazemi-etal-2023-lambada} or end-to-end~\cite{clark-etal-2020-transformers,picco-etal-2021-neural-unification} logical reasoning over small rulebases converted to NL. 
These approaches require both user-provided if/then rules to operate, 
while \sysname{} 
requires only facts and is thus applicable to domains in which rules are not available.

\paragraph{Modular Reasoning over NL}
\sysname{}'s systematic reasoning is related to approaches that decompose problems into sequences of modular operations.
\citet{gupta-etal-2020-neural} introduce a neural module network~[NMN;~\citet{andreas-etal-2016-neural}] for QA with modules for span extraction and arithmetic operations. \citet{khot-etal-2021-text} introduce another NMN that decomposes questions into simpler ones answerable by existing models. 
These are part of a broader class of work decomposing multi-step reasoning using reasoning modules~\cite{khot-etal-2022-decomposed,saha-etal-2023-branch-solve-merge,saha-etal-2023-murmur}.

\paragraph{Systematic Explanation Generation}
Recent works have used LMs to generate NL reasoning graphs in support of an answer.
``Chain-of-Thought'' prompting~\cite{wei-etal-2022-chain,kojima-etal-2022-large}, elicits free-form inference hops from the LM before it generates an answer.
Other text graph generation methods connect model-generated statements via common sense relations~\cite{bosselut-etal-2021-dynamic,arabshahi-etal-2021-conversational} or for/against influence~\cite{madaan-etal-2021-think,jung-etal-2022-maieutic}, though these are not knowledge-grounded and don't address entailment (see Figure \ref{fig:approaches}). 

The EntailmentBank dataset~\cite{dalvi-etal-2021-explaining} has driven research towards the construction of \textit{explanation trees}, challenging models to produce stepwise entailment proofs of a statement using a set of provided support facts. This direction builds upon works on explainable reasoning that build graphs from KB-retrieved sets of support statements, but stop short of showing their role in logical entailment~\cite{pan-etal-2019-improving,jansen-ustalov-2019-textgraphs,yadav-etal-2019-quick,valentino-etal-2022-case}.
Components of our framework are related to concurrent approaches for entailment tree construction~\cite{bostrom-etal-2022-natural,hong-etal-2022-metgen,sprague-etal-2022-natural}. \citet{ribiero-etal-2022-entailment} also use iterative retrieval-conditioned generation, and \citet{yang-etal-2022-generating} also use entailment classifiers to filter proof steps. 
None of the above tree generation approaches consider this harder scenario of multiple-choice QA, opting instead to focus on the reconstruction task from support facts.  \citet{tafjord-etal-2022-entailer} do consider the harder task. They propose a backward chaining QA system, Entailer, that generates entailment trees (not grounded in human-verified facts) using a search grounded to the model's internal beliefs, complementary to \sysname{}'s use of guided and retrieval-conditioned generation to obtain knowledge grounding. 
As with \sysname{} (\S\ref{sec:generalization}), Entailer benefits from humans adding more useful statements to its available knowledge~\cite{dalvi-mishra-etal-2022-towards}.
The difference between systems is highlighted by the number of generations (i.e. search nodes) considered by the two algorithms: {while Entailer considers at most a couple dozen hypothesis decompositions, {\sysname{} must consider hundreds or thousands} to find one that is fully grounded.}

\paragraph{Faithful Complex LLM Reasoning}
Of the growing body of work on LLM-based ``Chain-of-X'' reasoning methods~\cite{xia-etal-2024-beyond}, a portion considers ways to improve faithfulness to underlying knowledge and reduce LLM hallucinations (see \citet{lyu-etal-2024-towards} for an overview of faithful explanation methods).
\sysname{} contributes to this space by providing a faithful-by-design algorithm that reasons based on logical hypothesis grounding.
Other recent methods include Rethinking with Retrieval~\cite{he-etal-2022-rethinking}, which reranks reasoning chains using a faithfulness score based on retrieved knowledge,
and the forward-chaining Faithful Reasoning algorithm~\cite{creswell-shanahan-2022-faithful}, which follows ``a beam search over reasoning traces'' to answer a question by iteratively adding deductive inferences to a context of starting facts. Theirs is a different style search to our backward-chaining \sysname{}, not relying upon NLI for verifying logical connectedness and not scaling to larger knowledge bases.
\section{Background}
A logical expert system proves a propositional query against a \textit{theory} comprised of facts and inference rules, generally given in the form of Horn clauses. 
Upon finding a rule whose head can \textit{unify} with the query, a depth-first backward chaining algorithm such as those used in Prolog solvers will perform variable substitution and then recursively attempt to prove the terms in the rule's \textit{body}.
For example, a disease classification system might prove query $\scalemath{.9}{\textsc{Contagious}(\textit{flu})}$ via facts  $\scalemath{.9}{\textsc{Contagious}(\textit{influenza})}$ and
$\scalemath{.9}{\textsc{OtherName}(\textit{flu}, \textit{influenza})}$, and conjunctive rule $\scalemath{.9}{\textsc{Contagious}(X) \Leftarrow \textsc{OtherName}(X, Y) \land}$ $\scalemath{.9}{\textsc{Contagious}(Y)}$. It does so by unifying $\scalemath{.9}{\textsc{Contagious}(\textit{flu})}$ with $\scalemath{.9}{\textsc{Contagious}(X)}$ and then recursively unifying the terms in the rule body with their matching facts.
Here, \textit{flu} is an object symbol, \scalemath{.9}{\textsc{Contagious}} is a predicate symbol, and $X$ is a variable.
 See \citet{russell-norvig-2010-artificial} for further details. %
\begin{figure*}[t!]
    \centering
    \includegraphics[width=\textwidth,trim={27mm 1mm 5mm 4mm},clip]{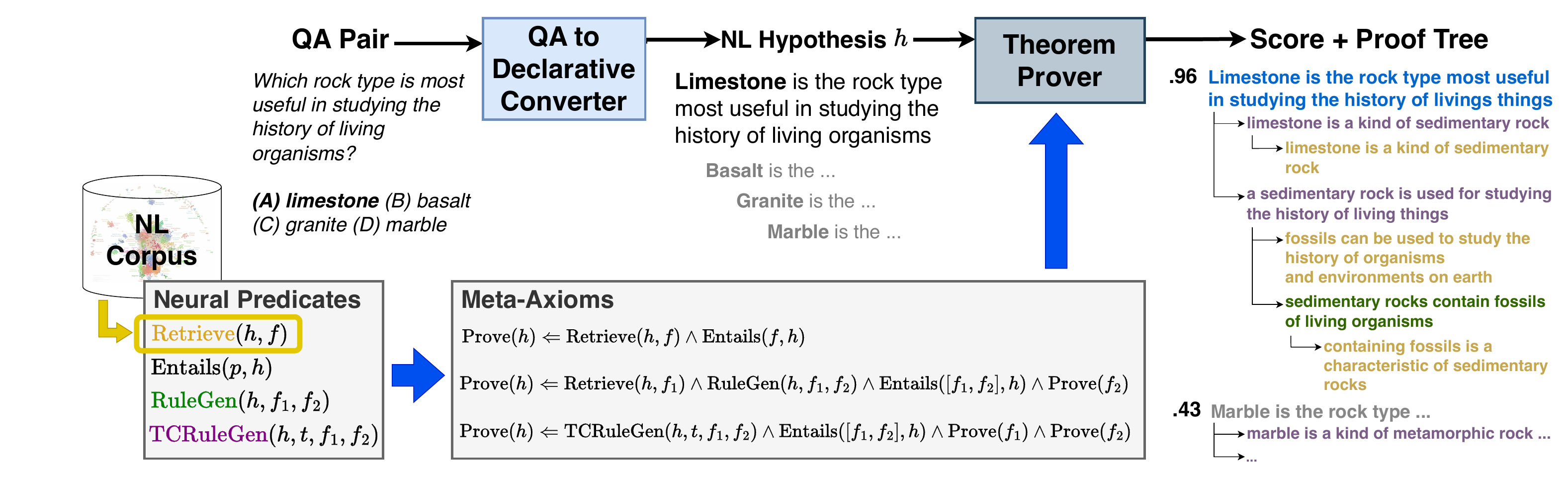}
    \caption{Proposed system framework. An off-the-shelf theorem prover searches for proofs of query $\textsc{Prove}(h)$, where symbol $h$ is an NL hypothesis translated from a QA pair. The prover uses a set of meta-axioms invoking neural retrieval, entailment, and generation predicates to dynamically instantiate inference rules that use the NL factbase.}
    \label{fig:overview}
\end{figure*}

\paragraph{Neural Predicates}
While most declarative predicates have meaning only in the context of user-defined inference axioms, others can
call external neural modules that produce values for their arguments or determine the truth value of the predicate. A popular implementation of this paradigm is DeepProbLog~\cite{manhaeve-etal-2018-deepproblog}, which we use to define LM-invoking predicates.
In the above example, we might train a sequence-to-sequence (seq2seq) model to produce other names for a disease $Y$, turning $\scalemath{.9}{\textsc{OtherName}(Y^+, X^-)}$\footnote{In Prolog syntax, `$\scalemath{.7}{+}$' denotes inputs, `$\scalemath{.7}{-}$' outputs.} into a neural predicate.
This mechanism creates the ability to introduce externally defined object symbols, e.g. seq2seq-generated NL, into the engine's vocabulary.

\paragraph{Weak Unification}
In classical backward chaining, a \textit{unification} operator assigns equivalence to two logical atoms; this requires atoms to have the same arity and have no unequal ground literals in the same argument position. Issues arise when literals are NL sentences, which can be syntactically distinct but semantically equivalent.
To handle this, \citet{weber-etal-2019-nlprolog} propose a \textit{weak unification} operator, which allows for the unification of \textit{any} same-arity atoms regardless of conflicting symbols.\footnote{Introducing weak unification greatly increases the search runtime, as one might try to unify any two symbols in the vocabulary at every recursive step. This poses a substantial challenge when applying the query grounding algorithm to a search space over NL.} 
They estimate a \textit{unification score} as the aggregation of pairwise similarity scores using a similarity function $\theta(s_1, s_2) \in [0,1]$.
The score of the full proof is taken as the minimum of scores across all steps.
In this work, we apply a similar aggregation; we say that a query fact $s_1$ ``weakly unifies'' with provenance fact $s_2$ with a unification score equal to the confidence of an NLI model taking $s_2$ as the premise and $s_1$ as the hypothesis.

\section{Overview of Approach}

Depicted in Figure \ref{fig:overview}, our framework is comprised of:
an external corpus of facts (some $\{f_1,\dots f_n\}$); a module that converts a QA pair to a hypothesis; an off-the-shelf theorem prover; and a suite of meta-axioms that use neural fact retrieval and dynamic rule generation modules to propose, verify, and score inferences. 
In our experiments, we consider one implementation of this framework that uses the corpus WorldTree~\cite{xie-etal-2020-worldtree}, a set of 9K  
NL science facts that can interchangeably be considered as rows in 81 $n$-ary relational tables.

\paragraph{Question Conversion}
Given a multiple-choice question, the system converts each candidate answer into a hypothesis $h$ using a Question to Declarative Sentence model~[QA2D; \citet{demszky-etal-2018-transforming}] (See \S
\ref{app:qa2d}).
It then searches for a proof of $h$ against its knowledge base.
For each alternative, we enumerate $p$ proofs using a time-capped backward chaining search and then take as the system's answer the candidate with the overall highest-scoring proof. 

\subsection{Inference Rule Structure}
Our approach uses LMs to dynamically generate inference rules given a hypothesis. The rule structure is strikingly simple, instantiating one of the following meta-level templates:
  \begin{enumerate}
      \item[I.] \texttt{Hypothesis} $\Leftarrow$ \texttt{Fact}
      \item[II.] \texttt{Hypothesis} $\Leftarrow$ \texttt{Fact1} $\land$ \texttt{Fact2}
  \end{enumerate} 
Via template I, the system proves the hypothesis by finding a provenance \texttt{Fact} stored in its knowledge store that entails the hypothesis. Via template II, it enumerates a pair, \texttt{Fact1} and \texttt{Fact2}, both either stored in the knowledge store or themselves recursively proved, such that the pair in conjunction entails the hypothesis.\footnote{We find that two-premise decomposition is sufficiently powerful and expressive for our purposes. For example, to prove a hypothesis such as those in EntailmentBank~\cite{dalvi-etal-2021-explaining} that requires a \textit{three}-premise conjunction $h \Leftarrow f_1 \land f_2 \land f_3$, \sysname{} produces instead a recursive set of decompositions $H \Leftarrow f_1 \land f_i; f_i \Leftarrow f_2 \land f_3$.
}
Template I is given higher search precedence than II,
yielding an intuitive high-level procedure:
we first look up the hypothesis against the factbase, searching for an entailing fact. If we do not find one, we decomposes the hypothesis into a pair of statements to be proved.
Concretely, for an input hypothesis $h$, we define the predicate $\textsc{Prove}(h)$ that serves as the primary goal term. 
We define the following core meta-rules, which use the neural predicates \textsc{Retrieve}, \textsc{Entails}, and \textsc{RuleGen}. At each step in the backward-chaining search, \sysname{}'s Prolog engine attempts to unify a query with the head of one of these three rules:
\begin{enumerate}[leftmargin=.6cm]
\item[1.] \textit{Fact Unification} \\ $\scalemath{.87}{\textsc{Prove}(h) \Leftarrow \textsc{Retrieve}(h^{+}, f^{-}) \land \textsc{Entails}(f, h)}$   
    
    \item[2.] \textit{Two Premise Rule Generation} \\ $\scalemath{.87}{\textsc{Prove}(h)\Leftarrow  \textsc{RuleGen}(h^+, f_1^-, f_2^-) \  \land} %
    \scalemath{.87}{\textsc{Entails}( [f_1, f_2], h)}$  \\ $\scalemath{.87}{\quad \land \ \textsc{Prove}(f_1) \land \textsc{Prove}(f_2)}$ 
    
    \item[3.] \textit{Retrieved First Premise Rule Generation} \\ $\scalemath{.87}{\textsc{Prove}(h)\Leftarrow 
    \textsc{Retrieve}(h^{+}, f_1^{-}) \ \land \ \textsc{RuleGen}(h^+, f_1^+, f_2^-)}  \\
    \scalemath{.87}{\quad \land \textsc{Entails}( [f_1, f_2], h)  \ \land \ } %
    \scalemath{.87}{\textsc{Prove}(f_2)}$
\end{enumerate}

\subsection{Unification with Retrieved Facts}
For factbase fact $f$, $\textsc{Prove}(f)$ is vacuously true.
Rule 1 shows how we prove $\textsc{Prove}(h)$ using retrieval.
The predicate $\textsc{Retrieve}$ proposes candidate $f_i$'s given $h$ using a FAISS~\cite{johnson-2019-billion}-based nearest neighbor dense retrieval index.
We train the retrieval encoder via ranking loss such that the embedding for a hypothesis is maximally similar to its supporting facts.
To promote logical coherence and improve the precision of the system, we apply a set of  \textbf{neural models for recognizing textual entailment (RTE)} as filters $\textsc{Entails}_j(\cdot)$ that iteratively rule out $f_i$ candidates that are not classified as entailing $h$. 
\[\scalemath{.87}{
\text{$\textsc{Entails}(f_i,h) \Leftarrow \bigwedge_{j =  1\dots n} \textsc{Entails}_j(f_i,h)$}}
\]
Implicit in rule 1 is that $\textsc{Prove}(h)$ weakly unifies with some $\textsc{Prove}(f_i)$; we assign the unification score $\theta(h, f_i)$ equal to the confidence of one RTE model. 

\begin{figure}
    \centering
    \includegraphics[width=\columnwidth,trim={1mm 1mm 2mm 1mm},clip]{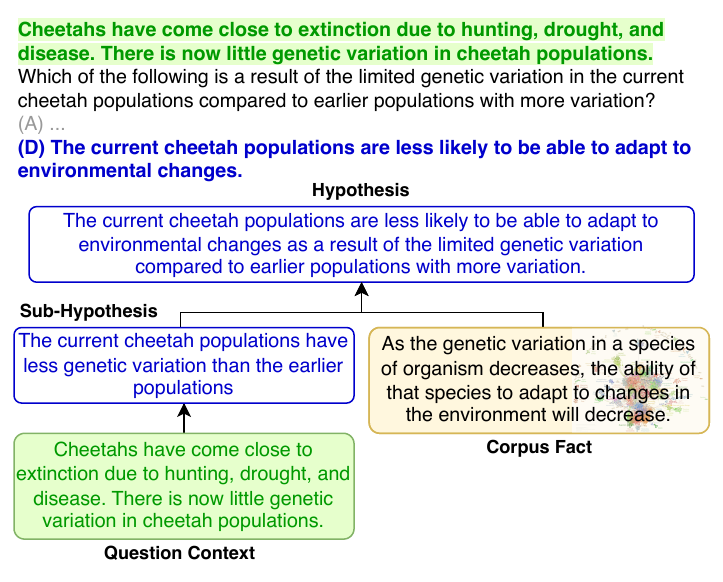}
    \caption{Example question in which a sub-hypothesis in the proof tree is grounded to the question context rather than to the factbase.}
    \label{fig:cheetah_grounding}
\end{figure}

For some questions such as the one depicted in \autoref{fig:cheetah_grounding}, it is necessary to ground a subquery in evidence from the problem.
To handle this, we add the question setup (defined as all but its last sentence) as a ``fact''  always proposed by $\textsc{Retrieve}$.

\subsection{Dynamic Rule Generator (DRG)}
If we do not find an entailing fact for hypothesis $h$, we decompose $h$ into a pair of entailing premises; we propose candidates using nucleus sampling~\cite{holtzman-etal-2019-curious} from a seq2seq model.
The predicate $\textsc{RuleGen}(h, f_1, f_2)$ prompts a model trained to generate $h \rightarrow f_1, f_2$ pairs. 
The space of potential decompositions is very large; there are many deductively valid ways to prove a hypothesis in natural language, though only a fraction of them are ultimately groundable in the provided factbase.
To bias the proof search towards those more likely to be grounded in the corpus, we adopt a two-pronged approach, illustrated in inference rules 2 and 3, that proposes a \textit{heterogeneous search frontier} using different biasing strategies in addition to straightforward LM sampling.

\paragraph{Template Conditioned Generation (TCG)}
One way in which we improve our LM-based proposal function is to leverage the high-level structure that supports reasoning in a given domain: we propose \textit{template-guided generation} to bias search towards the semi-structure of WorldTree.\footnote{While our approach is applicable to a wide array of reasoning problems, the use of WorldTree-specific templates illustrates one way by which we can infuse domain-specific structure into the neural search algorithm. 
This is a strict departure from the burdensome process of symbolic knowledge engineering.
As our experiments show, this guidance method improves \sysname{}'s QA accuracy by a few points, though ablating it yields a similarly strong performance.}
WorldTree tables correspond to types of facts with similar syntax and semantics that support scientific reasoning. 
WorldTree questions are annotated with the fact rows that support their answers.
Each table can be viewed as an \textit{n}-ary relation of columns whose values are text spans. E.g., the Taxonomic relation has columns 
\textit{\textless A\textgreater }, \textit{[HYPONYM]}, \textit{\textless is a / a kind of\textgreater }, \textit{\textless SCOPE1\textgreater },
\textit{[HYPERNYM]}, \textit{\textless for\textgreater }, \textit{[PURPOSE]}.
Rows include \textit{`{a bird is an animal}'} and \textit{`{a seed is a kind of food for birds}.'}

Thus, for the \textsc{RuleGen} predicate in rule 2, half of the $f_1 , f_2$ candidates are sampled from the DRG conditioned on $h$ plus a template that cues the model to reflect the syntax of a WorldTree table. 
We train the DRG to accept any masked infilling template (e.g. \textit{\textless mask\textgreater is a kind of \textless mask\textgreater }, akin to those used to pretrain LMs~\cite{lewis-etal-2019-bart,raffel-etal-2020-exploring}), and propose decompositions whose first fact reflects the template's syntax. We thus create a $h, t_1 \rightarrow f_1, f_2$ model.
We feed the model templates drawn from WorldTree's tables, {guiding it towards proof steps more likely to be grounded in the factbase.}
A sample of the 150 templates can be found in \autoref{fig:templates} (a larger list is shown in \S\ref{app:templates}).
We reuse the same model for non-template-conditioned generation by feeding it an empty template.
In practice, we make two generation calls: one samples $m$ free-generated candidates, and a second samples $n$ candidates for each of $n_t$ templates.
\begin{align*}
    \scalemath{.87}{\textsc{RuleGen}(h^+, f_1^-, f_2^-) \Leftarrow \textsc{Member} (t,  \textsc{Templates} \ \cup \{\texttt{BLANK}\} ) }   \\ 
    \scalemath{.85}{ \quad \land \ \textsc{TCRuleGen}(h^+, t^+, f_1^-, f_2^-)}
\end{align*}

\begin{figure}[t!]
    \centering
    \setlength{\tabcolsep}{1pt}
    \scriptsize
    \begin{tabular}{ll}
\toprule
                    \textbf{WorldTree Relation} &                                       \textbf{Template}  \\
\midrule
                      KINDOF &                     \textless mask\textgreater  is a kind of \textless mask\textgreater   \\
                      IFTHEN &                          if \textless mask\textgreater  then \textless mask\textgreater   \\
                 PROP-THINGS &                              \textless mask\textgreater  has \textless mask\textgreater   \\
                       CAUSE &                           \textless mask\textgreater  causes \textless mask\textgreater   \\
                      MADEOF &                          \textless mask\textgreater  made of \textless mask\textgreater   \\
                    REQUIRES &                         \textless mask\textgreater  requires \textless mask\textgreater   \\
                      ACTION &                          \textless mask\textgreater  is when \textless mask\textgreater   \\
                     USEDFOR &                          \textless mask\textgreater  used to \textless mask\textgreater   \\
                      PARTOF &                        \textless mask\textgreater  a part of \textless mask\textgreater   \\
                      CHANGE &                          \textless mask\textgreater  changes \textless mask\textgreater   \\
                     USEDFOR &                         \textless mask\textgreater  used for \textless mask\textgreater   \\
                      AFFECT &             \textless mask\textgreater  has \textless mask\textgreater  impact on \textless mask\textgreater   \\
          PROP-ANIMAL-ATTRIB &                      \textless mask\textgreater  is a \textless mask\textgreater  animal  \\
                    SOURCEOF &                   \textless mask\textgreater  is a source of \textless mask\textgreater   \\
                  COMPARISON &                             \textless mask\textgreater  than \textless mask\textgreater   \\
                    EXAMPLES &                 an example of \textless mask\textgreater  is \textless mask\textgreater   \\
         COUPLEDRELATIONSHIP &         as \textless mask\textgreater  decreases \textless mask\textgreater  will \textless mask\textgreater   \\
           PROP-CONDUCTIVITY &                     \textless mask\textgreater  is \textless mask\textgreater  conductor  \\
                PROP-GENERIC &                 \textless mask\textgreater  is a property of \textless mask\textgreater   \\
                     HABITAT &                          \textless mask\textgreater  live in \textless mask\textgreater   \\
                MEASUREMENTS &                  \textless mask\textgreater  is a measure of \textless mask\textgreater   \\
\bottomrule
\end{tabular}
    \caption{Sample of WorldTree templates used for guided generation.}
    \label{fig:templates}
\end{figure}

\paragraph{Template Selection}
WorldTree is a diverse corpus, containing tables that are specific to a particular subset of science problems (e.g. the ``Predator-Prey'' table). As conditioning on dozens of templates can be computationally expensive, we introduce a \textit{case-based reasoning}~\cite{schank-1983-dynamic,das-etal-2021-case} approach that selects relevant templates for a given hypothesis. We construct an Okapi-BM25~\cite{jones-etal-2000} retrieval index over {questions} from the WorldTree QA training set to obtain
the most lexically similar items to a query. At inference time, we select the top-$k$ most similar questions to the query and take as our template subset the tables of the questions' annotated facts.

\paragraph{Retrieval Conditioned Generation (RCG)}
In rule 3, rather than generate a pair of subqueries, we \textit{immediately ground} half of the antecedent by choosing as $f_1$ a fact retrieved directly from the corpus. We have the DRG {force decode} $f_1$ before generating $f_2$ as normal, then recur only on $f_2$.

\paragraph{Filters} %
As stochastic sampling from LMs can be noisy, some fraction of the generated candidate set may be invalid: premises may be incoherent, or the decomposition might not properly entail $h$.
Accordingly, we introduce a set of compositional entailment verifiers~\cite{khot-etal-2020-qasc,jhamtani-clark-2020-learning} trained on two-premise compositional entailment.
We also add a ``self-ask'' filter, which following \citet{tafjord-etal-2022-entailer} is an LM fine-tuned to assign a statement a truth value. If the confidence is below 0.5 for either an entailment judgment or the `self-ask' belief in $f_1$ or $f_2$, then we filter the pair.
All filters condition on the question text as context.
When \sysname{} uses these rules, the unification score equals the lowest of the scores for $\textsc{Prove}(f_1)$, for $\textsc{Prove}(f_2)$, and the confidence of entailment filters $s_e([f_1, f_2] \Rightarrow H)$.

\subsection{Proof Search}
Given a query, \sysname{} searches for $t$ seconds to find up to $p$ proofs of depth $d$ or less. We follow \citet{weber-etal-2019-nlprolog} in pruning search branches whose unification score is guaranteed to fall below the current best, given our monotonic aggregation function $\min(\cdot)$. Found proofs that score under the current best do not count towards $p$.
Full pseudocode for the algorithm, which follows a depth-first search with a breadth-first lookahead~\cite{stern-etal-2010-using} to check for the unification of generated subgoals, can be found in \S\ref{app:algo}.
It is parameterized by 
\begin{enumerate}[itemsep=0pt]
    \item A maximum number of proofs $m$ at which to cut off searching. In experiments, we set this to $10$ for top-level queries and $2$ for recursive subqueries.
    \item A number of support facts $n_f$ to retrieve at each call to $\textproc{Retrieve}_K$, which we set to 15.
    \item Candidate generation rates $n_v$ for vanilla nucleus-sampled decompositions, $n_t$ for template-conditioned decompositions, and $n_r$ for retrieval-conditioned generations. We set these each to $40$.\footnote{Due to batching, these correspond to 3 total calls to the HuggingFace library's \texttt{generate} function for the T5-3B model.}
    Upon removing exact match duplicates, a call to $\textproc{RuleGen}$ produces about $100$ candidates.
    \item Entailment scoring module $\textsc{Score}_e(\cdot)$, which is a separate RTE cross-encoder model for single- and double-premise entailments.
\end{enumerate}

\section{Experiments}
We train the components of \sysname{} to be able to answer questions in the Science QA domain. The different neural modules are trained on reformulations of existing datasets for scientific reasoning. Further information can be found in \S\ref{app:model-training}.

Our experiments illustrate how the approach exemplified by \sysname{} provides grounded and logical explanations, performing comparably or better than approaches that do not satisfy these properties.
We evaluate models on two multiple-choice QA datasets constructed so that correct answers are supported by facts in the WorldTree corpus:

\noindent\textbf{EntailmentBank}~\cite{dalvi-etal-2021-explaining} is a dataset of entailment trees for declarativized answers to the AI2 Reasoning Challenge (ARC) dataset~\cite{clark-etal-2018-think}, showing how the hypothesis can be derived via compositional entailment hops from WorldTree facts.
We recast the test set, initially designed to test tree reconstruction, into QA by retrieving the corresponding multiple-choice ARC questions 
from which the hypotheses were constructed. 

\noindent\textbf{WorldTree}~\cite{xie-etal-2020-worldtree} is a subset of the ARC dataset annotated with undirected explanation graphs whose nodes are facts from the WorldTree tablestore. 
We note that WorldTree explanations do not show how facts should combine. \textit{There is no guarantee that fully grounded trees exist for these questions using the WorldTree corpus alone}. 
The difficulty of this task is akin to that of a course exam: the teacher (us) provides the student (the model) with a very large study guide of facts, but expects the student to \textbf{use} these facts by composing them to reason coherently about a problem.

Our task metric is {accuracy}: whether a generated proof of the correct option outscores any other.\footnote{If it produces no proofs, it gives no answer and is wrong. 
\sysname{} searches for up to $p$=10 proofs of max depth $d$=5 with a timeout of $t$=180 seconds per option. 
}

\paragraph{Baselines}
We  evaluate \nellie{} against \textbf{Entailer}~\cite{tafjord-etal-2022-entailer}, another system that produces entailment tree proofs via backward chaining. Entailer stops recurring \textbf{not} when it finds entailing facts from a corpus, but rather when \textit{the model believes} that a subquery is true with high confidence. Its trees are thus {not} grounded in verified facts.
We reimplemented their algorithm using their T5-11B-based model, reporting their configuration of a max tree depth (\textbf{D}) of 3 and minimum of 1 (i.e. $\geq1$ decomposition). 
We also evaluate their ablated setting which recurs exactly once (D=1). This is a baseline that is neither grounded nor particularly interpretable, generating just a pair of statements.
To isolate the impact of model size, we also evaluate an \textbf{Entailer-3B} model based on the same T5-3B model as \sysname{}'s DRG.\footnote{We obtained the training data from \citet{tafjord-etal-2022-entailer} and trained our model in consultation with the authors.} 
We also compare 
against grounded xQA methods without logical structure (see Fig \ref{fig:approaches}): \textbf{PathNet}~\cite{kundu-etal-2019-exploiting}, which constructs 2-hop chains by linking entities between facts, 
and three approaches that build graphs from facts using linear programming: \textbf{TupleILP}~\cite{khot-etal-2017-answering}, \textbf{ExplanationLP}~\cite{thayaparan-etal-2021-explainable}, and \textbf{\textit{Diff}-Explainer}~\cite{thayaparan-etal-2022-diff}.\footnote{We show results from [\citet{thayaparan-etal-2021-explainable} \& \citeyear{thayaparan-etal-2022-diff}].}

\begin{table}[t]
\setlength{\tabcolsep}{5pt}
\def\arraystretch{1.2}%
\footnotesize
\centering
\begin{tabular}[b]{lccrrr}
\toprule 
&  \multicolumn{2}{c}{\textbf{Explanations}} & \multicolumn{3}{c}{\textbf{QA Accuracy (\%)}} \\ \cmidrule(lr){2-3} \cmidrule(lr){4-6}
& \textbf{Grounded} & \textbf{Logical} & \textbf{Ovr} & \textbf{Easy} & \textbf{Chal} \\  

\midrule
\multicolumn{6}{l}{\textbf{EntailmentBank QA}} \\ \midrule
\textbf{\sysname{} (3B)} & \yes & \yes & 71.4 & 76.4 & 60.4 \\
Entailer-3B \\
\quad (D $=$ 3)  & \no & \yes & 48.7 & 52.8 & 39.6 \\
\quad (D $=$ 1) &  \no & \no  & 64.9 & 71.2 & 50.9  \\ \hline %
Entailer-11B  \\
\quad (D $=$ 3) & \no & \yes & 73.2 & 77.3 & 64.2 \\
\quad (D $=$ 1) & \no & \no & 71.1 & 76.4 & 59.4 \\% & \no & \yes
\midrule 
\multicolumn{6}{l}{\textbf{WorldTree QA}} \\ \midrule
\textbf{\sysname{} (3B)}   & \yes & \yes & 71.4 & 75.7 & 60.9 \\
Entailer-3B  \\
\quad (D $=$ 3) & \no & \yes & 45.2 & 50.1 & 33.3 \\
\quad (D $=$ 1)  & \no & \no & 47.7 & 51.6 & 38.5 \\  \hline
Entailer-11B \\
\quad (D $=$ 3) & \no & \yes & 73.2 & 76.7 & 64.6 \\
\quad (D $=$ 1)  & \no & \no & 74.1 & 78.7 & 63.0 \\  \hline
PathNet & \yes & \no & 43.4 &  &  \\
TupleILP & \yes & \no & 49.8 & \\
ExplanationLP & \yes & \no & 62.6 &  &  \\
Diff-Explainer & \yes & \no & 71.5 &  &  \\
\bottomrule
\end{tabular}
\caption{\sysname{} performance vs. comparable XQA systems.
}
\label{tab:qa_results}
\end{table}

\subsection{Results}
Table \ref{tab:qa_results} shows QA performance.
\sysname{}'s overall (\textbf{Ovr}) accuracy of over 70\% 
{matches that of the best-performing grounded baseline}, \textit{Diff}-Explainer, on WorldTree, and is {within 2 points of the best logically directed baseline}, Entailer-11B. 
This is notable given Entailer-11B is a much larger model and has no requirement to provide grounded proofs. \sysname{} {outperforms the same-sized Entailer-3B baseline} by a margin of 22\% on EntailmentBank and 26\% on WorldTree under its default max D = 3 configuration, and even outperforms the less explainable D = 1 variant. 

\begin{figure}[t!]
    \centering
    \includegraphics[width=.9\columnwidth]{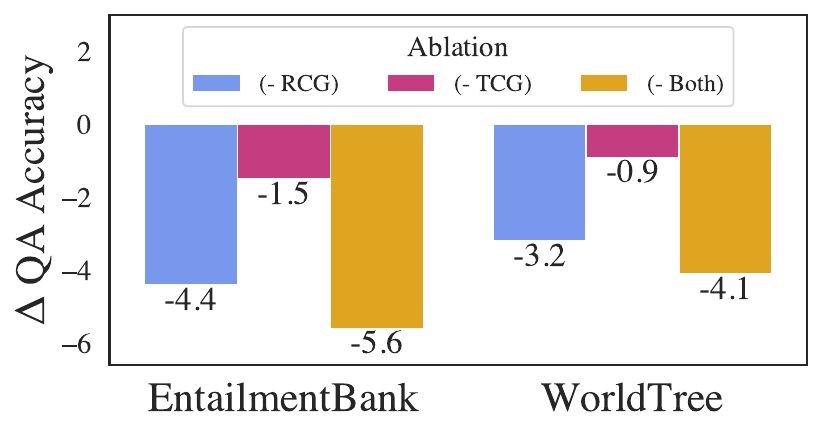}
    \captionsetup{skip=5pt}
    \caption{Effect of ablating one or both of rule-conditioned (RCG) and template-conditioned generation (TCG).}
    \label{fig:ablations}
\end{figure}

Figure \ref{fig:ablations} shows the results of ablating \sysname{}'s two conditional generation modules and replacing them with vanilla generation.
{Removing both template- and retrieval-conditioned generation lowers \sysname{}'s performance in all circumstances}, highlighting the empirical benefit of structured guidance. 
Ablating TCG (($-$ TCG), and the drop from ($-$ RCG) to ($-$ Both)) reduces performance by 1-1.5 points.
Ablating RCG (\sysname{} vs ($-$ RCG)) drops it by 3-4.

\subsection{Domain Generalization and Knowledge Scaling}
\label{sec:generalization}
While \sysname{} was trained on datasets centered around WorldTree, we show that it can perform in another domain by altering the knowledge over which it reasons. We consider OpenBookQA~\cite{mihaylov-etal-2018-suit}, a dataset that requires reasoning over facts not included in WorldTree; each question is associated with one WorldTree science fact (``metals conduct electricity''), but also one fact from a separate pool of common knowledge (``a suit of armor is made of metal'').

\begin{figure}[t!]%
\centering
\includegraphics[width=.49\columnwidth]{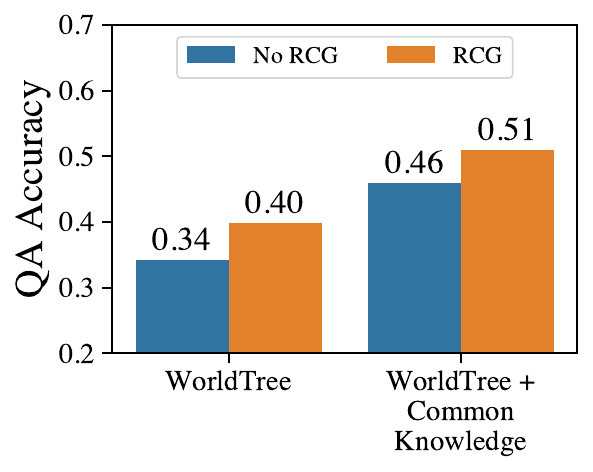}
\includegraphics[width=.49\columnwidth]{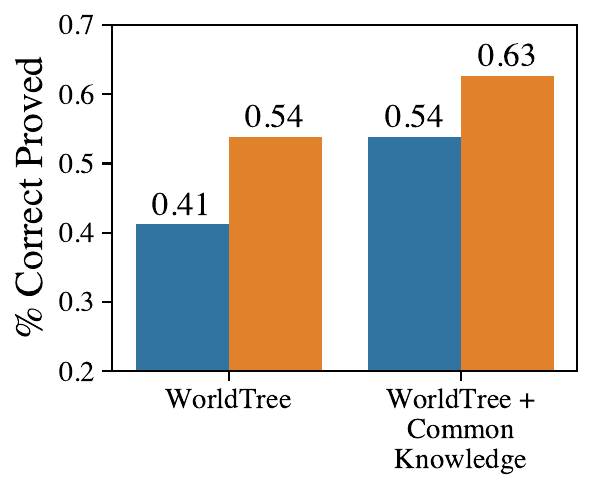}
\captionsetup{skip=5pt}
\caption{\sysname{} QA accuracy and proof recall on OBQA with vs. without access to common knowledge statements}
\label{fig:obqa}
\end{figure}

\begin{figure*}[htb!]
    \includegraphics[width=\textwidth]{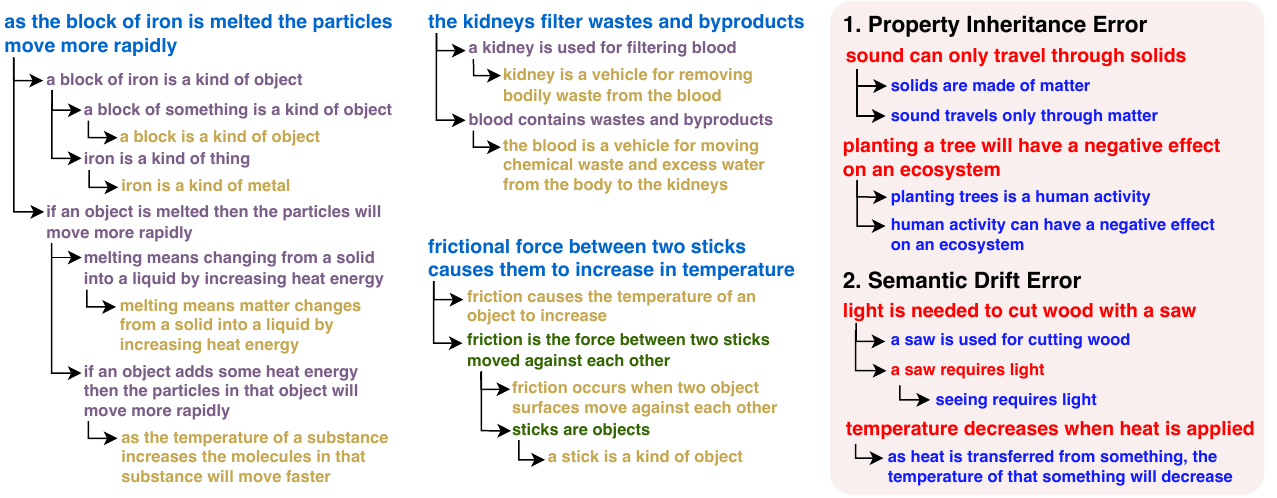}
\caption{\textbf{(Left)} Example \sysname{} proofs. \textcolor{dblue2}{\textbf{Top-level queries}} are decomposed into subqueries via \textcolor{dgreen}{\textbf{retrieval-}} or \textcolor{dpurple}{\textbf{template-}}conditioned generation. Proof leaves are \textcolor{dorange}{\textbf{corpus facts}}.
\textbf{(Right)} Common classes of error causing \textcolor{red}{\textbf{false}} statements to be grounded in \textcolor{blue}{\textbf{true}} ones.} 
\label{fig:proofs}
\end{figure*}

\begin{figure}[t!]
    \centering
    \includegraphics[width=.3\textwidth]{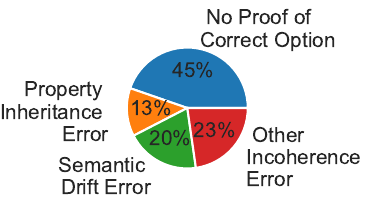}
    \captionsetup{skip=5pt}
    \caption{Distribution of \sysname{} error classes among 136 incorrect answers on EntailmentBank questions}
    \label{fig:errors_pie}
\end{figure}

As with a classical expert system, we have designed \sysname{} to be ``improvable merely by making statements to it''~\cite{mccarthy-1959-programs}. This suggests that as we increase the provided knowledge store, we should expect \sysname{} to reason more effectively. Because WorldTree does not cover all the knowledge required to answer OBQA questions, we test whether \sysname{} performs better on them when we add the common knowledge to its retrieval index. As the common knowledge annotations are one fact per question and are not designed specifically for entailment, it is unlikely a priori that a {fully} grounded entailment tree can be created using the available knowledge alone, making this a challenging task.

Figure \ref{fig:obqa} shows \sysname{}'s accuracy with and without RCG. We do not use TCG, as the targeted common knowledge is free-form. 
We find that QA accuracy increases 11-12\% with the common knowledge added. The percent of correct statement proved also increases 9-12\%. These results also highlight the importance of RCG, as it provides a 5\% QA boost and 13\% on correct proof rate.
These results show that \sysname{} can be applied out-of-the-box to a dataset requiring reasoning over a different source of free-form NL knowledge.

\subsection{Tree Error Analysis}
\label{sec:qualitative}
We find that \sysname{} can produce high quality proofs of correct hypotheses; examples are shown in Figure \ref{fig:proofs} (left) and 
\S F. 
However, the system can also generate proofs for incorrect answers that bypass its entailment filters. We can diagnose 
(and perhaps annotate) 
error patterns to address in future work. We list a few categories, depicted in Figure \ref{fig:proofs} (right).
A common pattern of error is \textbf{Hypernym Property Inheritance Errors} (also known as the ``fallacy of the undistributed middle''),
 in which the model assumes that a member of a taxonomic class has a property of their hypernym, but the property is not universal. A colloquial example is inferring \textit{penguins can fly} from \textit{penguins are birds} \& \textit{birds can fly}.
We also observe an amount of \textbf{Errors from Semantic Drift}~\cite{khashabi-etal-2019-possibilities} between inference hops, culminating in RTE model false positives. E.g., in block $2$ of Fig \ref{fig:proofs} (right), which object loses temperature during heat transfer changes between hops. Figure \ref{fig:errors_pie} shows the distribution of these errors on a set of 136 incorrect answers from EntailmentBank. The predominant error case is a failure to find any proof of the correct option.

\paragraph{Correct Answer Tree Analysis}
To investigate whether \sysname{} reaches a right solution with right vs wrong reasoning, we
manually inspected for reasoning errors in 50 correct answer
trees produced by NELLIE. We found that 39/50 (78\%) of trees
were perfectly acceptable. Most of the 11 unacceptable trees
were due to incompleteness at one decomposition step.

\section{Conclusion}
We propose a reimagined version of a classical expert system that relies on the inferential power of LLMs rather than handcrafted rules. \sysname{} has the skeleton of a symbolic theorem prover, but the provided rules invoke neural predicates that allow for systematic reasoning over NL statements. 
To dynamically generate inferences, we introduce two mechanisms for knowledge-guided generation: conditioning decompositions on retrieval or model-selected inference templates. 
Our search algorithm undergirds \sysname{}, an explainable reasoning system that performs end-to-end QA by searching for grounded entailment trees.
We find that \sysname{} equals or exceeds the performance of comparable XQA systems, while providing explanations with the simultaneous guarantees of logical directionality and groundedness in human-provided text. 
This work thus suggests a new way to jointly reap the benefits of both modern neural
methods and traditional reasoning.

\section*{Acknowledgements}
Thanks to Nils Holzenberger, Kyle Richardson, Elias Stengel-Eskin, Marc Marone, Kate Sanders, Rachel Wicks, and Andrew Blair-Stanek for comments on earlier drafts. 
Thank you to Abbey Coogan for feedback on figure design.

\bibliographystyle{named}
\bibliography{anthology_plus_custom}

\appendix
\newpage
\section{Model Details and Training}
\label{app:model-training}
We train the components of \sysname{} to be able to answer questions in the Science QA domain. The different neural modules are trained on reformulations of existing datasets for scientific reasoning.
We use the following sources:
\myparagraph{WorldTree Explanations~\cite{xie-etal-2020-worldtree}} and
\myparagraph{EntailmentBank~\cite{dalvi-etal-2021-explaining}} are used as described in the paper.
\myparagraph{eQASC~\cite{jhamtani-clark-2020-learning}} and \myparagraph{SciTaiL~\cite{khot-etal-2018-scitail}} are datasets of two- and one-premise entailment questions. 

\myparagraph{QA2D}
Our QA2D model is a T5-11B~\cite{raffel-etal-2020-exploring} fine-tuned by \citet{tafjord-etal-2022-entailer} on the original QA2D dataset~\cite{demszky-etal-2018-transforming}.

\paragraph{Retrieval Module}
The dense retrieval model is a {SentenceTransformers} Siamese BERT-Network encoder~\cite{reimers-gurevych-2019-sentence} pretrained on the MS MARCO IR corpus~\cite{nguyen-etal-2016-ms}, which we fine-tune via ranking loss to maximize the cosine similarity between a hypothesis and {its support facts}.
We gather $h,f_1$ from the WorldTree, EntailmentBank, and eQASC training sets, then sample random negatives during training.

\paragraph{Dynamic Rule Generator}
The DRG is a T5-3B model~\cite{raffel-etal-2020-exploring} fine-tuned on $h \rightarrow [f_1, f_2]$ pairs drawn from eQASC, which are numerous (26K) but noisy,
and the binary composition steps from the EntailmentBank training set, which are few (3.8K) but ultimately rooted in facts from WorldTree.
To make the DRG able to condition on templates, we adopt the span masking strategy from \citet{raffel-etal-2020-exploring} to create a randomly masked version $\hat{f}_1$ of each $f_1$ and then train the model on $h, \hat{f}_1 \rightarrow f_1, f_2$ pairs. For each original triplet, we create one training example with a template, and one in which $\hat{f}_1$ is an empty \textless mask\textgreater , giving the model the dual capacity for non-template-generation.

\paragraph{Entailment Filters}
We find that a mixture of strategies helps prioritize precision over recall when filtering candidates.
For 1-premise filtering for unification and 2-premise filtering for rule generation, we fine-tune a separate pair of models consisting of a {SentenceTransformer} cross-encoder classifier and a T5 seq2seq. 
The former is trained via classification loss, the latter via sequence cross entropy loss.
The 1-premise filters are trained on the SciTail dataset, the 2-premise filters on eQASC and EntailmentBank.
For 2-premise entailment, we additionally add a ``self-ask'' filter, which following \citet{tafjord-etal-2022-entailer} is a T5 model fine-tuned to produce its 0-to-1 confidence that a statement is true. If the confidence is below 0.5 for either $f_1$ or $f_2$, we filter the pair. As in \citet{tafjord-etal-2022-entailer}, we train our DRG seq2seq in a multitask fashion so that it serves as both the rule generator as well as the ``self-ask'' filter and (half of) the 2-premise entailment filter.

\subsection{Training}
We use the default training parameters from the SentenceTransformers library\footnote{\url{https://tinyurl.com/mvhbmfcc}} in order to train CrossEncoder-based classifier and Bi-encoder retriever models for 3 epochs. 
We correspondingly use the default fine-tuning parameters from the HuggingFace library in order to train our T5-based generation and classification models.

\section{Hyperparameters}
\label{app:hparams}
The DRG produces $40$ (pre-filter) candidates divided evenly amongst the set of templates drawn via case-based retrieval from the top-($k$=10) most similar questions to a given query hypothesis. It also generates $40$ free-generated candidates, and $40$ retrieval-conditioned ones ($4$ each for the $10$ top-scoring retrieved $f_1$'s). It therefore considers $120$ candidate decompositions at each recursion step. All sampling is done via nucleus sampling ($p=.95$). 
RTE models filter all candidates for which the classification softmax likelihood of entailment is less than $0.7$. We remove duplicates before filtering.

We used Optuna~\cite{akiba-etal-2019-optuna} to find the number of facts for retrieval-conditioned generation, the $p$ value for nucleus sampling, the entailment threshold, and the number of candidates generated at each recursion step. We used as our objective the overall QA accuracy on the EntailmentBank dev set. We ran it for about 300 trials, and found that the most important parameters were the entailment threshold (too low = too many bad proofs, too high = no proofs) and the nucleus $p$ (less than 0.8 reduced QA performance by 10-15. Experiments were run on 24GB Quadro RTX 6000s.

\begin{figure}[t!]
    \centering
    \setlength{\tabcolsep}{1pt}
    \scriptsize
    \begin{tabular}{ll}
\toprule
                    \textbf{WorldTree Relation} &                                       \textbf{Template}  \\
\midrule
                   INSTANCES &                               \textless mask\textgreater  is \textless mask\textgreater   \\
                      ACTION &                               \textless mask\textgreater  to \textless mask\textgreater   \\
                      KINDOF &                     \textless mask\textgreater  is a kind of \textless mask\textgreater   \\
                 PROP-THINGS &                              \textless mask\textgreater  are \textless mask\textgreater   \\
                      ACTION &                              \textless mask\textgreater  for \textless mask\textgreater   \\
                 AFFORDANCES &                              \textless mask\textgreater  can \textless mask\textgreater   \\
                      IFTHEN &                          if \textless mask\textgreater  then \textless mask\textgreater   \\
                 PROP-THINGS &                              \textless mask\textgreater  has \textless mask\textgreater   \\
                       CAUSE &                           \textless mask\textgreater  causes \textless mask\textgreater   \\
                      MADEOF &                          \textless mask\textgreater  made of \textless mask\textgreater   \\
                 PROP-THINGS &                             \textless mask\textgreater  have \textless mask\textgreater   \\
                    REQUIRES &                         \textless mask\textgreater  requires \textless mask\textgreater   \\
                      ACTION &                          \textless mask\textgreater  is when \textless mask\textgreater   \\
                     USEDFOR &                          \textless mask\textgreater  used to \textless mask\textgreater   \\
                      PARTOF &                        \textless mask\textgreater  a part of \textless mask\textgreater   \\
                      CHANGE &                          \textless mask\textgreater  changes \textless mask\textgreater   \\
                     USEDFOR &                         \textless mask\textgreater  used for \textless mask\textgreater   \\
       ATTRIBUTE-VALUE-RANGE &                            \textless mask\textgreater  means \textless mask\textgreater   \\
                      AFFECT &             \textless mask\textgreater  has \textless mask\textgreater  impact on \textless mask\textgreater   \\
          PROP-ANIMAL-ATTRIB &                      \textless mask\textgreater  is a \textless mask\textgreater  animal  \\
                    SOURCEOF &                   \textless mask\textgreater  is a source of \textless mask\textgreater   \\
                    CONTAINS &                         \textless mask\textgreater  contains \textless mask\textgreater   \\
         COUPLEDRELATIONSHIP &                          as \textless mask\textgreater  will \textless mask\textgreater   \\
                  CHANGE-VEC &                        \textless mask\textgreater  decreases \textless mask\textgreater   \\
                  COMPARISON &                             \textless mask\textgreater  than \textless mask\textgreater   \\
                      ACTION &                \textless mask\textgreater  is when \textless mask\textgreater  to \textless mask\textgreater   \\
                       CAUSE &                        \textless mask\textgreater  can cause \textless mask\textgreater   \\
                   LOCATIONS &                            \textless mask\textgreater  found \textless mask\textgreater   \\
                  COMPARISON &                             \textless mask\textgreater  more \textless mask\textgreater   \\
       PROP-INHERITEDLEARNED &                \textless mask\textgreater  is \textless mask\textgreater  characteristic  \\
                      DURING &                           \textless mask\textgreater  during \textless mask\textgreater   \\
                  CHANGE-VEC &                        \textless mask\textgreater  increases \textless mask\textgreater   \\
                 AFFORDANCES &                    \textless mask\textgreater  can \textless mask\textgreater  by \textless mask\textgreater   \\
                    EXAMPLES &                 an example of \textless mask\textgreater  is \textless mask\textgreater   \\
                    SOURCEOF &                         \textless mask\textgreater  produces \textless mask\textgreater   \\
                    CONTAINS &                          \textless mask\textgreater  contain \textless mask\textgreater   \\
                      CHANGE &                         \textless mask\textgreater  converts \textless mask\textgreater   \\
                  COMPARISON &                   \textless mask\textgreater  is \textless mask\textgreater  than \textless mask\textgreater   \\
                    FORMEDBY &                        \textless mask\textgreater  formed by \textless mask\textgreater   \\
            CONSUMERS-EATING &                              \textless mask\textgreater  eat \textless mask\textgreater   \\
                    SOURCEOF &                         \textless mask\textgreater  provides \textless mask\textgreater   \\
    PROP-RESOURCES-RENEWABLE &                      \textless mask\textgreater  is \textless mask\textgreater  resource  \\
       ATTRIBUTE-VALUE-RANGE &                  \textless mask\textgreater  means \textless mask\textgreater  of \textless mask\textgreater   \\
                  COMPARISON &                             \textless mask\textgreater  less \textless mask\textgreater   \\
         COUPLEDRELATIONSHIP &         as \textless mask\textgreater  decreases \textless mask\textgreater  will \textless mask\textgreater   \\
           PROP-CONDUCTIVITY &                     \textless mask\textgreater  is \textless mask\textgreater  conductor  \\
                PROP-GENERIC &                 \textless mask\textgreater  is a property of \textless mask\textgreater   \\
                     HABITAT &                          \textless mask\textgreater  live in \textless mask\textgreater   \\
                MEASUREMENTS &                  \textless mask\textgreater  is a measure of \textless mask\textgreater   \\
         COUPLEDRELATIONSHIP &         as \textless mask\textgreater  increases \textless mask\textgreater  will \textless mask\textgreater   \\
\bottomrule
\end{tabular}
    \caption{Partial list of WorldTree templates used for guided generation, sorted by frequency of occurrence in EntailmentBank trees (full list omitted for space).}
    \label{tab:template-list}
\end{figure}

\section{QA2D Conversion}
\label{app:qa2d}
The following is an example conversion by our QA2D model of a multiple-choice question into a set of candidate hypotheses to prove.

\textbf{Q: Ethanol is an alternative fuel made from corn. What is one of the unfavorable effects of using ethanol as a fuel?}
\begin{enumerate}[itemsep=0pt]
    \item[(A)] decreased cost of fuel production
    \item[(B)] decrease in farm land available for food production
    \item[(C)] increase in the consumption of fossil fuels
    \item[(D)] increased carbon footprint from driving automobiles
\end{enumerate}

Converted to the following hypotheses:
\begin{enumerate}[itemsep=0pt]
    \item[(A)] Using ethanol has an unfavorable effect on the cost of fuel.
    \item[(B)] Using ethanol as fuel decreases farm land available for food production.
    \item[(C)] Using ethanol as fuel increases in the consumption of fossil fuels.
    \item[(D)] The use of ethanol as an alternative fuel can lead to increased carbon footprints.
\end{enumerate}

\section{WorldTree Templates}
\label{app:templates}
Table \ref{tab:template-list} depicts the infilling templates gathered from WorldTree and used for template-guided rule generation. We manually annotated this list. The full set of templates is of size $150$.

\section{Search Algorithm} 
\label{app:algo}
Algorithm \ref{alg:backward-chaining} depicts \sysname{}'s search algorithm, the logic of which is implemented in a meta-interpreter coded in DeepProbLog~\cite{manhaeve-etal-2018-deepproblog}. 

\section{Example Proofs}
\begin{figure*}[t!]
\centering
\footnotesize
\begin{tabular}{m{5cm} m{11cm}}
\toprule 
Question & \sysname{} Proof \\ \midrule 
In what way can forest fires affect the lithosphere? \teal{(A) increasing soil erosion} (B) causing air pollution (C) increasing seismic activity (D) causing violent dust storms & \includegraphics[width=10.5cm]{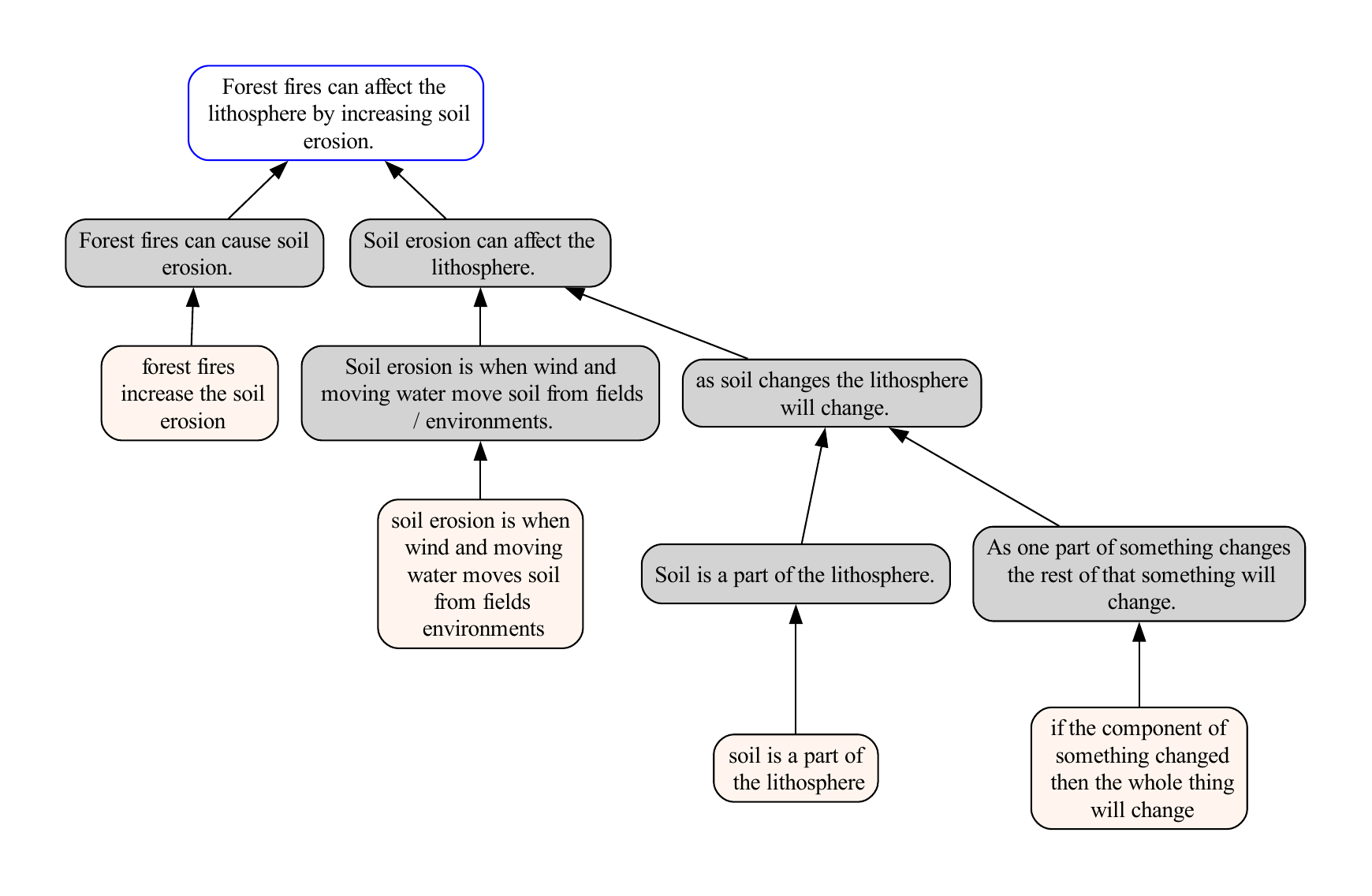} \\
The snowshoe hare sheds its fur twice a year. In the summer, the fur of the hare is brown. In the winter, the fur is white. Which of these statements best explains the advantage of shedding fur? (A) Shedding fur keeps the hare clean. (B) Shedding fur helps the hare move quickly. (C) Shedding fur keeps the hare's home warm. \teal{(D) Shedding fur helps the hare blend into its habitat.} & \includegraphics[width=10.5cm]{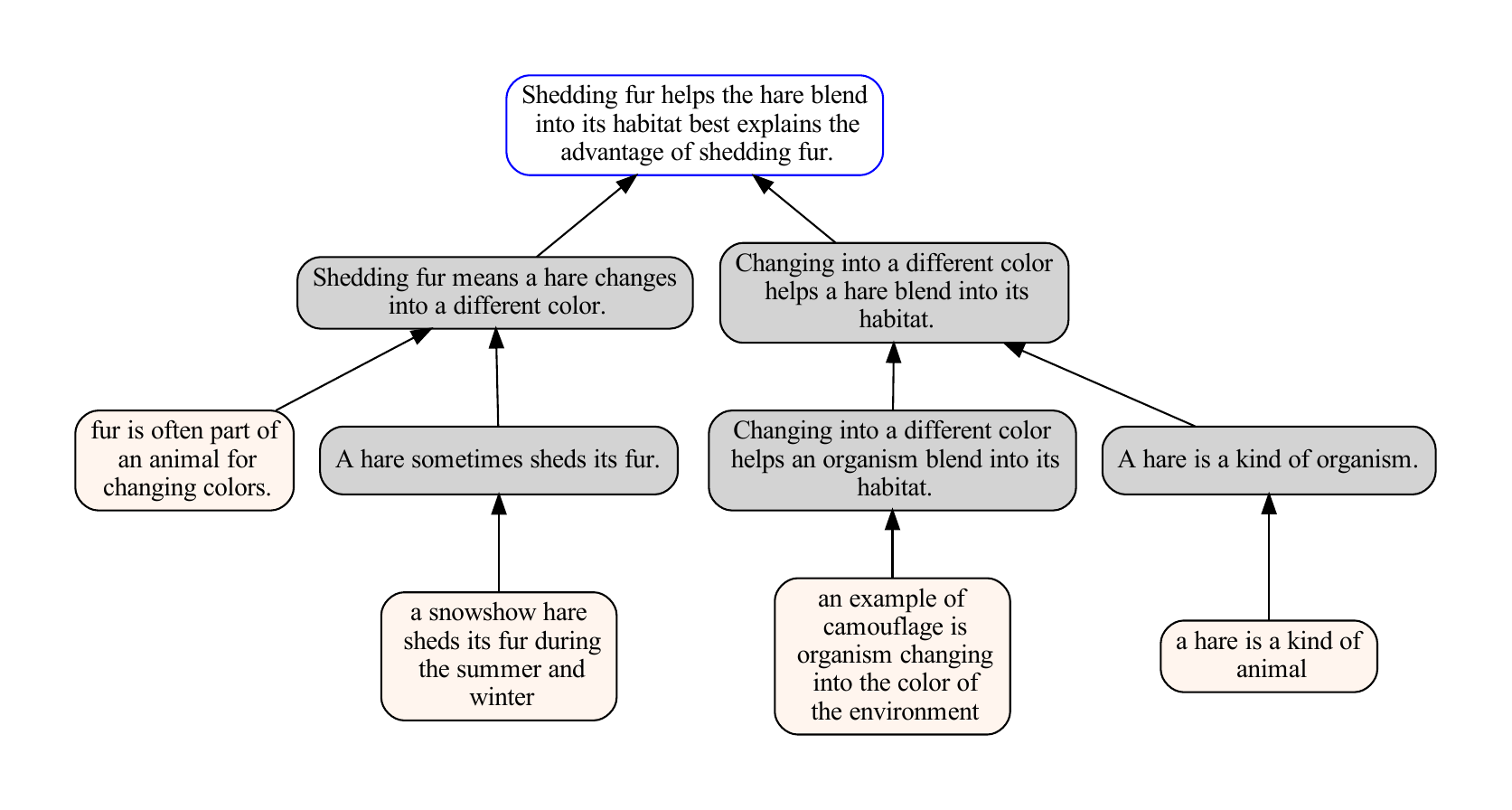} \\
Scientists use large optical telescopes to obtain information about the planets in the solar system. What wavelengths of electromagnetic radiation provide this information? (A) gamma radiation (B) infrared radiation (C) radio waves \teal{(D) visible light} & \includegraphics[width=10.5cm]{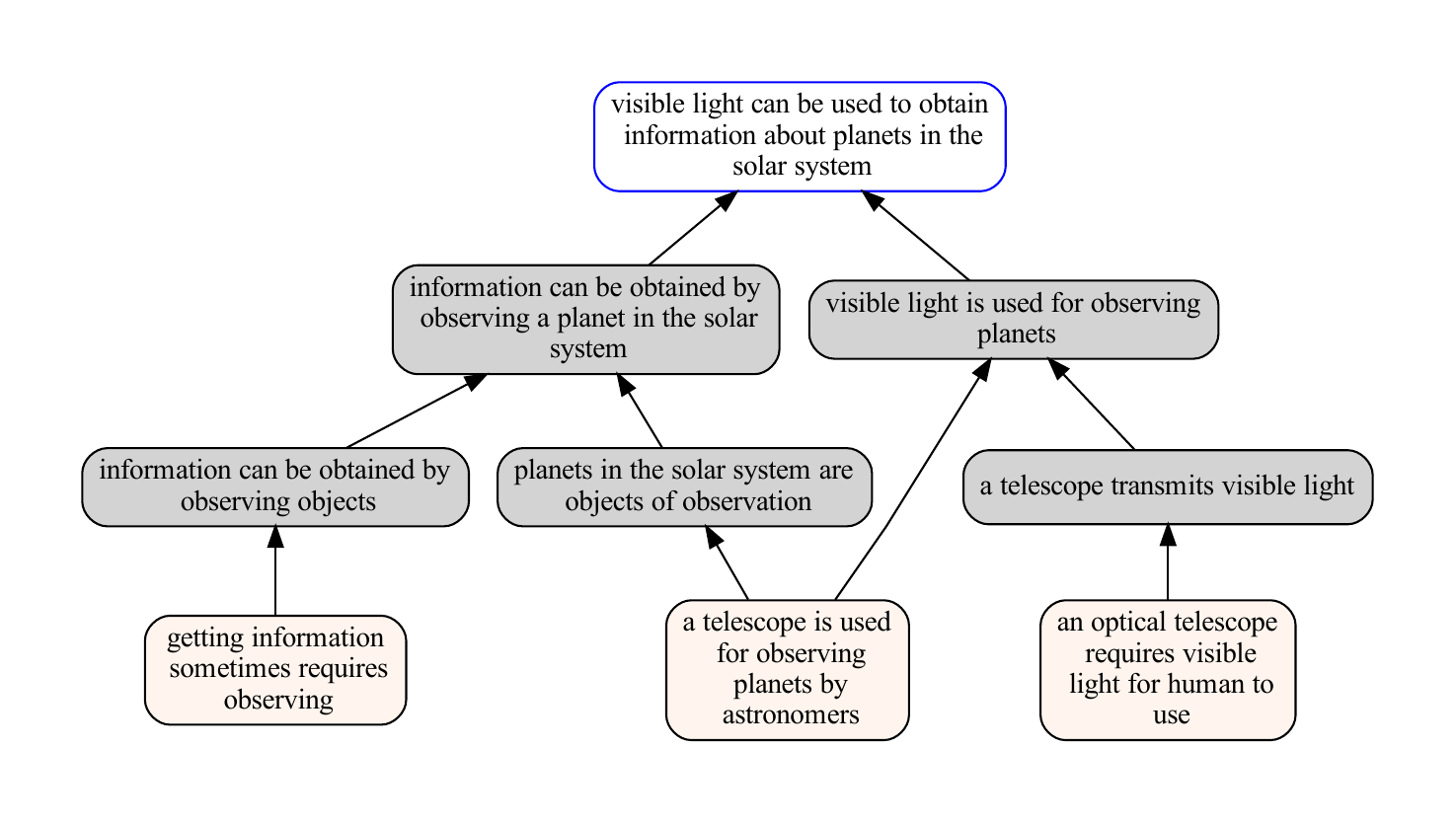} \\
\bottomrule
\end{tabular}
\caption{Example multiple-choice questions with NELLIE's answer and corresponding proof. Blue-outlined nodes are hypotheses generated by the QA2D model for the correct answer highted in \teal{green}. Grey nodes are generated subqueries; tan nodes are leaves from \sysname{}'s knowledge corpus.}
\label{fig:more-trees}
\end{figure*}

Figure \ref{fig:more-trees} illustrates more proofs generated by \sysname{} for questions from EntailmentBank and WorldTree.

\section{Impact of Design Choices}
\sysname{} is comprised of numerous modules that are very difficult to evaluate in isolation given a lack of in-domain test sets for tasks like QA2D, decomposition generation ($H \rightarrow F1,F2$) compositional RTE ($H \Leftarrow F1,F2$), single-premise RTE ($H \Leftarrow F1$), and fact retrieval. Below, we enumerate some modeling choices, alternatives we tried, and the extent to which we observed them impact the system.

\begin{itemize}
    \item \textbf{QA2D:}\\
    \textbf{\sysname{} uses:} \citet{tafjord-etal-2022-entailer}'s T5-11B model.\\
    \textbf{We also tried:} using GPT-J~\cite{wang-komatsuzaki-2021-gpt-j} with in-context-learning exemplars drawn from EntailmentBank. This results in \textbf{10+ point drop in QA}, mainly because (A) EntailmentBank hypotheses are designed to be \textit{generics} instead of \textit{question-conditioned statements} and (B) GPT-J tended to generate true statements even for wrong answer options. 
    \item \textbf{Decomposition generation ($H \rightarrow F1,F2$):} \\
    \textbf{\sysname{} uses:} A T5-3B model trained in a multi-angle fashion similar to \citet{tafjord-etal-2022-entailer}, using additional data for compositional entailment and decomposition generation. \\
    \textbf{We also tried:} Using a T5-Large model trained only on the decomposition data. This allowed us to generate {far more decomposition candidates per hypothesis}, but rate of obtaining a \textit{good} decomposition was far lower. This led to \textit{5+ point drop in QA}. 
    \item \textbf{Compositional RTE ($H \Leftarrow F1,F2$):}\\
    \textbf{\sysname{} uses:} A sequence of (1) The same T5-3B model as for generation and (2) a cross-encoder trained on eQASC and EntailmentBank. \\
    \textbf{We also tried:} (A) using a bi-encoder trained via contrastive loss (this was not as effective at ruling out bad decompositions), (B) not training the bi-encoder to condition on the question, and (C) only training on EntailmentBank (as \citet{tafjord-etal-2022-entailer} do). All three variants
    reduce QA by 3-7\%.
    \item \textbf{Single-premise RTE ($H \Leftarrow F1$):}\\
    \textbf{\sysname{} uses:} (1) A cross-encoder trained on Sci-Tail\\
    \textbf{We also tried:} an off-the-shelf RTE model\footnote{\url{https://huggingface.co/cross-encoder/nli-deberta-v3-base}}, which had too high a false positive rate to be effective. 
    \item \textbf{Fact Retrieval:}\\
    \textbf{\sysname{} uses:} a Sentence-Transformers bi-encoder pretrained on MS MARCO and fine-tuned on $(H, F1)$ pairs drawn from WorldTree, EntailmentBank, and eQASC.
    \\
    \textbf{We also tried:} the same bi-encoder off-the-shelf without finetuning. This only dropped performance by 3-4 points.
    
\end{itemize}

\algrenewcommand\algorithmicindent{1.0em}%
\algrenewcommand\algorithmiccomment[1]{\hfill {\it // #1}}

\begin{algorithm*}[ht!]
\begin{algorithmic}[1]
\small{
\Procedure{\textcolor{blue}{Prove}}{$H$, $m$, $d$} \ra{} proofs $tree^{(1)}(H), \dots tree^{(m)}(H)$ \& scores $s^{(1)}(H),\dots s^{(m')}{(H)}$:
\State i. {\bf check weak unification} against $K$:
\State \gapxx         entailing premises $P = \textcolor{blue}{\textproc{PremiseLookup}}(H)$ 
\State \gapxx         \textbf{if} $P$ nonempty
\State \gapxx \textbf{then return} proofs$=[(p \vdash{} H) \text{ for } p \in P]$, scores$=[\textcolor{orange}{\textproc{Score}_e}(p \vdash{} H) \text{ for } p \in P]$ 
\State \gapxx \textbf{elif} $d = d_{\text{max}}$
\State \gapxx \textbf{then Fail} 
\State ii. {\bf generate candidate decompositions:} \Comment{If $H$ does not unify against $K$, decompose it}
\State \gapxx  decompositions $D = \textcolor{blue}{\textproc{RuleGen}}(H)$
\State iii. {\bf check whether any decompositions pairs both unify:} \Comment{this step amortizes premise lookups}
\State \gapxx proofs = []; scores = []
\State \gapxx  \textbf{forall} $(p_1, p_2) \in D$ \textbf{do}
\State \gapxxxxxx  \textbf{if} $p_1 \in K$ \ \textbf{and} \ $\exists p_2' \in \textcolor{blue}{\textproc{PremiseLookup}}(p_2)$ \Comment{if $p_1$ already grounded, its subtree is leaf}
\State \gapxxxxxx \textbf{then} proofs += $(\{p_1, (p_2' \vdash{} p_2)\} \vdash{} H)$
\State \gapxxxxxxxx scores += $\min(\textcolor{orange}{\textproc{Score}_e}(p_2'\vdash{} p_2), \textcolor{orange}{\textproc{Score}_e}(p_1 \land p_2 \vdash{} H))$ \Comment{tree score is min of scores used to construct it}
\State \gapxxxxxx  \textbf{elif} $\exists p_1' \in \textcolor{blue}{\textproc{PremiseLookup}}(p_1)$ \ \textbf{and} \ $\exists p_2' \in \textcolor{blue}{\textproc{PremiseLookup}}(p_2)$ 
\State \gapxxxxxx \textbf{then} proofs += $(\{(p_1' \vdash{} p_1), (p_2' \vdash{} p_2)\} \vdash{} H)$
\State \gapxxxxxxxx scores += $\min(\textcolor{orange}{\textproc{Score}_e}(p_1'\vdash{} p_1), \textcolor{orange}{\textproc{Score}_e}(p_2'\vdash{} p_2), \textcolor{orange}{\textproc{Score}_e}(p_1 \land p_2 \vdash{} H))$
\State  \gapxx \textbf{if} proofs not empty
\State  \gapxx \textbf{then} \textbf{return} proofs, scores
\State iv. \textbf{recur on generated premises:} \Comment{if still no proof, need to recur}
\State \gapxx proofs = []; scores = []
\State \gapxx \textbf{forall} $(p_1, p_2) \in D$ \textbf{do}
\State \gapxxxxxx \textbf{if} $\textcolor{orange}{\textproc{Score}_e}(p_1\land p_2 \vdash{} H) > \max(\text{scores})$ \ \textbf{and} \ $\text{len}(\text{proofs}) < m$ \Comment{only recur if chance of better score}
\State \gapxxxxxx \textbf{then} $(\text{proof}_{p_1},S_{p_1}) = \textcolor{blue}{\textproc{Prove}}(p_1, 1, d+1)$  
\State \gapxxxxxxxx $(\text{proof}_{p_2},S_{p_2}) = \textcolor{blue}{\textproc{Prove}}(p_2, 1, d+1)$ 
\State \gapxxxxxxxx recursive score $S_r = \min(S_{p_1}, S_{p_2}, \textcolor{orange}{\textproc{Score}_e}(p_1 \land p_2 \vdash{} H))$
\State \gapxxxxxxxx \textbf{if} $S_r > \max(\text{scores})$
\State \gapxxxxxxxx \textbf{then }proofs += $(\{\text{proof}_{p_1}, \text{proof}_{p_2}\} \vdash{} H)$
\State \gapxxxxxxxxxx scores += $S_r$ 
\State \gapxx \textbf{return} proofs, scores
\EndProcedure
\State 
\Procedure{\textcolor{blue}{PremiseLookup}}{H} \ra{} premises $P$ s.t. $\forall p \in P, p \vdash{} H$ \Comment{try to ground H in K via weak entailment}
\State \gapxx         candidate premises $P = \textcolor{orange}{\textproc{Retrieve}_{K}}(H, n_f)$ \Comment{retrieve candidate premises s.t. $p \vdash{} H$}
\State \gapxx \textbf{forall} $f \in \textcolor{orange}{\textproc{OnePremiseFilters}}$ \textbf{do}
\State \gapxxxxxx         $P = f(P, H)$  \Comment{only keep premises that pass all filters}
\State \gapxx \textbf{return} $P$
\EndProcedure
\State
\Procedure{\textcolor{blue}{RuleGen}}{$H$} \ra{}  decompositions $D = (p_1^1, p_2^1), \dots (p_1^c, p_2^c)$ \Comment{find candidate pairs s.t. $p_1 \land p_2 \vdash{} H $}
\State \gapxx $D$ = []
\State \gapxx $D$ += $\textcolor{orange}{\textproc{Generate}}(H, n = n_v)$ 
\State \gapxx \textcolor{dgreen}{\textproc{WTTemplates}} = \textcolor{orange}{\textproc{CaseBasedRetrieveTemplate}}(H)
\State \gapxx outputs per template $o_t = n_t / \text{len}(\textcolor{dgreen}{\textproc{WTTemplates}})$
\State \gapxx \textbf{forall} $t \in \textcolor{dgreen}{\textproc{WTTemplates}}$ \textbf{do} \Comment{generate candidates for each of 50 templates in \textproc{WTTemplates} } 
\State \gapxxxxxx $D$ += $\textcolor{orange}{\textproc{Generate}}([H \ t], n = o_t)$  \Comment{T5 model accepts optional $t$ appended to $H$}
\State \gapxx retrieved first premises $P_1 = \textcolor{orange}{\textproc{Retrieve}_{K}}(H, n_f)$
\State \gapxx outputs per premise $o_r = n_r / \text{len}(P_1)$
\State \gapxx \textbf{forall} $p_{K} \in P_1$ \textbf{do} \Comment{generate candidates with retrieved support facts as first premise}
\State \gapxxxxxx $D$ +=  $\textcolor{orange}{\textproc{Generate}}(H, \text{prefix} = p_K, n=o_r)$   \Comment{Force decode $p_K$ then generate a $p_2$}
\State \gapxx \textbf{forall} $f \in \textcolor{orange}{\textproc{TwoPremiseFilters}}$ \textbf{do} \Comment{includes entailment, regex, self-asking filters}
\State \gapxxxxxx   $D = f(D, H)$  \Comment{only keep premise pairs that pass all filters}
\State \gapxx  \textbf{return} $D$
\EndProcedure

}
\end{algorithmic}

\caption{\sysname{}'s backchaining algorithm, coded in Prolog, for searching for proof list $tree^{(1)}(H), \dots tree^{(m')}(H)$ with scores $s^{(1)}(H),\dots s^{(m')}(H); m' \leq m$ for a hypothesis $H$ against factbase $K$. The rule generator produces $n_v$ candidates via vanilla nucleus sampling, $n_t$ candidates via sampling from templates, and $n_r$ candidates whose first fact is retrieved from K. The search is capped at max depth $d_{\text{max}}$.
All \textcolor{orange}{\textproc{Generate}}, \textcolor{orange}{\textproc{Retrieve}}, and entailment scorer \textcolor{orange}{$\textproc{Score}_e$} calls in \textbf{forall} loops are batched. All \textproc{module} outputs are cached. \textcolor{orange}{\textproc{OnePremiseFilters}} and \textcolor{orange}{\textproc{TwoPremiseFilters}}
\ are ordered lists of neural entailment models. $\vdash{}$ denotes weak entailment. \label{alg:backward-chaining}}
\end{algorithm*}

\end{document}